# Memoization of Top-down Parsing

*Mark Johnson*
*Cognitive and Linguistic Sciences, Box 1978*
*Brown University*
Mark_Johnson@Brown.edu

## 1. Introduction

In a paper published in this journal, Norvig (1991) pointed out that memoization of a top-down recognizer program produces a program that behaves similiarly to a chart parser. This is not surprising to anyone familiar with logic-programming approaches to NLP. For example, the Earley deduction proof procedure is essentially a memoizing version of the top-down SLD proof procedure employed by Prolog. Pereira and Warren (1983) showed that the steps of the Earley Deduction proof procedure proving the well-formedness of a string $S$ from the standard 'top-down' DCG axiomatization of a CFG $G$ correspond directly to those of Earley's algorithm recognizing $S$ using $G$.

Yet as Norvig notes in passing, using his approach the resulting parsers in general fail to terminate on left-recursive grammars, even with memoization. The goal of this paper is to discover why this is the case and present a functional formalization of memoized top-down parsing for which this is not so. Specifically, I show how to formulate top-down parsers in a 'continuation-passing style' which incrementally enumerates the right string positions of a category, rather than returning a set of such positions as a single value. This permits a type of memoization not described to my knowledge in the context of functional programming before. This kind of memoization is akin to that used in logic programming, and yields terminating parsers even in the face of left recursion.

In this paper, algorithms are expressed in the Scheme programming language (Rees and Clinger 1991). Scheme was chosen because it is a popular, widely known language which many readers find easy to understand. Scheme's 'first-class' treatment of functions simplifies the functional abstraction used in this paper, but the basic approach can be implemented in more conventional languages as well. Admittedly elegance is a matter of taste, but personally I find the functional specification of CFGs described here as simple and elegant as the more widely known logical (DCG) formalization, and I hope that the presentation of working code will encourage readers to experiment with the ideas described here and in more substantial works such as Leermakers (1993). In fact, my own observations suggest that with minor modifications (such as the use of integers rather than lists to indicate string positions, and vectors indexed by string positions rather than lists in the memoization routines) an extremely efficient chart parser can be obtained from the code presented here.

Ideas related to the ones discussed here have been presented on numerous occasions. Almost 20 years ago Shiel (1976) noticed the relationship between chart parsing and top-down parsing. Leermakers (1993) presents a more abstract discussion of the functional treatment of parsing, and avoids the left-recursion problem for memoized functional parsers by using a 'recursive ascent' or PLR parsing strategy instead of a top-down strategy. At a more abstract level than that of this paper, Shieber, Schabes and Pereira (1994) show that a variety of well-known parsing algorithms can be viewed as computing the closure of a set of basic parsing operations on a representation of the input string.

## 2. Formalizing context-free grammars

It is fairly straight-forward to implement a top-down parser in a functional programming language. The key insight is that a non-terminal category $A$ in a grammar defines a function $f_A$ that maps a string position $l$ in the input string $\gamma$ to a set of string positions $f_A(l)$ such that $r \in f_A(l)$ iff $A$ can derive the substring of $\gamma$ spanning string positions $l$ to $r$ (see e.g., Leermakers (1993) for discussion).

For example, suppose `V`, `NP` and `S` are already bound to $f_V$, $f_{NP}$ and $f_S$, and the grammar contains the following productions with VP on the left hand side.

(1) VP → V NP     VP → V S

Then the following Scheme definition binds `VP` to $f_{VP}$.

(2)
```
(define (VP p)
  (union (reduce union ’() (map NP (V p)))
         (reduce union ’() (map S (V p)))))
```

If sets are represented by unordered lists, `union` can be given the following definition. The function `reduce` is defined such that an expression of the form (`reduce` $f\, e$ '$(x_1 \ldots x_n)$) evaluates to $(f\,( \ldots (f\, e\, x_1) \ldots )\, x_n)$.

(3)
```
(define (reduce fn init args)
  (if (null? args)
      init
      (reduce fn (fn init (car args))
                 (cdr args))))
```

(4)
```
(define (union set1 set2)
  (if (null? set1)
      set2
      (if (member (car set1) set2)
          (union (cdr set1) set2)
          (cons (car set1)
                (union (cdr set1) set2)))))
```

When evaluated using Scheme's applicative-order reduction rule, such a system behaves as a depth-first, top-down recognizer in which non-determinism is simulated by backtracking. For example, in (2) the sequence V NP is first investigated as a potential analysis of VP, and then the sequence V S is investigated.

Rather than defining the functions *f* by hand as in (2), higher-order functions can be introduced to automate this task. It is convenient to use suffixes of the input string to represent the string positions of the input string (as in DCGs).

The expression (`terminal` *x*) evaluates to a function which maps a string position *l* to the singleton set { *r* } iff the terminal *x* spans from *l* to *r*, and the empty set otherwise.

(5)
```
(define (terminal X)
  (lambda (p)
    (if (and (pair? p)
             (eq? (car p) X))
        (list (cdr p))
        '())))
```

The expression (`seq` $f_A$ $f_B$) evaluates to a function which maps a string position *l* to the set of string positions { $r_i$ } such that there exists an $m \in f_A(l)$, and $r_i \in f_B(m)$. Informally, the resulting function recognizes substrings that are the concatenation of a substring recognized by $f_A$ and a substring recognized by $f_B$.

(6)
```
(define (seq A B)
  (lambda (p)
    (reduce union '() (map B (A p)))))
```

The expression (`alt` $f_A$ $f_B$) evaluates to a function which maps a string position *l* to $f_A(l) \cup f_B(l)$. Informally, the resulting function recognizes the union of the substrings recognized by $f_A$ and $f_B$.

(7)
```
(define (alt A B)
  (lambda (p)
    (union (A p) (B p))))
```

While `terminal`, `seq` and `alt` suffice to define (epsilon-free) context-free grammars, we can easily define other useful higher-order functions. For example, `epsilon` recognizes the empty string (i.e., it maps every string position *l* into the singleton set { *l* }), (`opt` $f_A$) recognizes an optional constituent, and (`k*` $f_A$) recognizes zero or more occurences of the substrings recognized by $f_A$.

(8)
```
(define epsilon list)
```

(9)
```
(define (opt A) (alt epsilon A))
```

(10)
```
(define (k* A)
  (alt epsilon
       (seq A (k* A))))
```

These higher-order function can be used to provide simpler definitions, such as (2a) or (2b), for the function `VP` defined in (2) above.

```
(2a)    (define VP (alt (seq V NP) (seq V S)))
```

```
(2b)    (define VP (seq V (alt NP S)))
```

This method of defining the functions corresponding to categories is quite appealing. Unfortunately, most Scheme implementations are deficient in that they do not allow mutually recursive functional definitions of the kind in (2a) or (2b). For example, suppose `S` is defined as in (11) and `VP` is defined as in (2a).

```
(11)    (define S (seq NP VP))
```

Further suppose (11) precedes (2a) textually in the program. Then the variable `VP` in (11) will be incorrectly interpreted as unbound. Changing the order of the definitions will not help, as then the variable `S` will be unbound.[1] A work-around is to add a vacuous lambda abstraction and application as in (11a), in effect delaying the evaluation of function definition.

```
(11a)   (define S (lambda args (apply (seq NP VP) args)))
```

With a macro definition such as (12) (named to remind us of this deficiency of current Scheme implementations and perhaps encourage implementors to do better in the future), the definition of functions such as (11a) can be written as (11b).

```
(12)    (define-syntax vacuous
          (syntax-rules ()
            ((vacuous fn)
             (lambda args (apply fn args)))))
```

```
(11b)   (define S (vacuous (seq NP VP)))
```

Figure 1 contains a fragment defined in this way. After these definitions have been loaded, an expression such the one in (13) can be evaluated. It returns a list of the input string's suffixes that correspond to the right string position of an S.

```
(13)    > (s '(Kim knows every student likes Sandy))
        ((likes sandy) ())
```

In example (13), the list resulting from the evaluation contains two suffixes, corresponding to the fact that both *Kim knows every student* and *Kim knows every student likes Sandy* can be analysed as Ss.

Finally, the `recognize` predicate can be defined as follows. The expression (`recognize` *words*) is true iff *words* is a list of words that can be analysed as an S, i.e., if

---

1 This problem can arise even if syntactic constructions specifically designed to express mutual recursion are used, such as `letrec`. Although these variables are closed over, their values are not applied when the defining expressions are evaluated, so such definitions should not be problematic for an applicative-order evaluator. Apparently most Scheme implementions require that mutually recursive functional expressions *syntactically* contain a `lambda` expression. Note that this is not a question of reduction strategy (e.g., normal-order versus applicative-order), but an issue about the syntactic scope of variables.

```
(define S (vacuous (seq NP VP)))                  ; S → NP VP
(define VP (vacuous (alt (seq V NP)               ; VP → V NP
                         (seq (V S)))))           ;    | V S
(define NP (vacuous (alt PN                       ; NP → PN
                         (seq Det N))))           ;    | Det N
(define PN (alt (terminal 'Kim) (terminal 'Sandy)))
(define V (alt (terminal 'likes) (terminal 'knows)))
(define Det (alt (terminal 'every) (terminal 'no)))
(define N (alt (terminal 'student) (terminal 'professor)))
```

**Figure 1: A CFG fragment defined using the higher-order constructors**

the empty string is a one of right string positions of an S whose left string position is the whole string to be recognized.

(14)
```
(define (recognize words)
  (member ’() (S words)))
```

## 3. Memoization and left recursion

As noted above, the Scheme function defined in this way behave as top-down, backtracking recognizers. It is well-known that such parsing methods suffer from two major problems.

First, a top-down parser using a left-recursive grammar typically fails to terminate on some inputs. This is true for recognizers defined in the manner just described; left-recursive grammars yield programs which contain ill-founded recursive definitions.[2]

Second, backtracking parsers typically involve a significant amount of redundant computation, and parsing time is exponential in the length of the input string in the worst case. Again, this is also true for the recognizers just described.

Memoization is a standard technique for avoiding redundant computation, and as Norvig (1991) noted, it can be applied to top-down recognizers to convert exponential-time recognizers into polynomial-time recognizers.

A general way of doing this is by defining a higher-order procedure `memo` which takes a function as an argument and returns a memoized version of it.[3] This procedure is essentially the same as the `memoize` predicate that is extensively discussed in Abelson and Sussman (1985).

---

2 Specifically, if *A* is a Scheme variable bound to the function corresponding to a left-recursive category, then for any string position *p* the expression (*A p*) reduces to another expression containing (*A p*). Thus the (applicative-order) reduction of such expressions does not terminate.

3 For simplicity, the `memo` procedure presented in (15) stores the memo table as an association list, in general resulting in a less than optimal implementation. As Norvig notes, more specialized data structures, such as hash tables, can improve performance. In the parsing context here, optimal performance would probably be obtained by encoding string positions with integers, allowing memo table look-up to be a single array reference.

(15)
```
(define (memo fn)
  (let ((alist '()))
    (lambda args
      (let ((entry (assoc args alist)))
         (if entry
             (cdr entry)
             (let ((result (apply fn args)))
               (set! alist (cons (cons args result)
                                 alist))
               result))))))
```

To memoize the recognizer, the original definitions of the functions should be replaced with their memoized counterparts; e.g., (11b) should be replaced with (11c). Clearly these definitions could be further simplified with suitable macro definitions or other 'syntactic sugar'.

(11c)
```
(define S (memo (vacuous (seq NP VP))))
```

As an aside, it is interesting to note that memoization can be applied selectively in this approach. For example, because of the overhead of table lookup in complex feature based grammars it might be more efficient not to memoize all categories, but rather restrict memoization to particular categories such as NP and S.

Now we turn to the problem of left recursion. In a logic programming setting, memoization (specifically, the use of Earley deduction) avoids the non-termination problems associated with left-recursion, even when used with the DCG axiomatization of a left-recursive grammar. But as Norvig mentions in passing, with parsers defined in the manner just described, the memoized versions of programs derived from left-recursive grammars fail to terminate.

It is easy to see why. A `memo`-ed procedure constructs an entry in a memo table only *after* the result of applying the unmemoized function to its arguments has been computed. Thus in cases of left recursion, memoization does nothing to prevent the ill-founded recursion that leads to non-termination.

In fact it is not clear how memoization could help in these cases, given that we require that `memo` behaves semantically as the identity function; i.e., that (`memo` *f*) and *f* are the same function. Of course, we could try to weaken this identity requirement (e.g., by only requiring that (*f x*) and ((`memo` *f*) *x*) are identical when the reduction of the former terminates), but it is not clear how to do this systematically.

Procedurally speaking, it seems as if memoization is applying 'too late' in the left-recursive cases; reasoning by analogy with Earley deduction, we need to construct an entry in the memo table when such a function is *called*; not when the result of its evaluation is known. Of course, in the left-recursive cases this seems to lead to an inconsistency, since these are cases where the *value* of an expression is required to compute that very value.

Readers familiar with Abelson and Sussman (1985) will know that in many cases it is possible to circumvent such apparent circularity by using asynchronous 'lazy streams' in place of the list representations (of string positions) used above above. The continuation-passing style encoding of CFGs discussed in the next section can be seen as a more functionally oriented instantiation of this kind of approach.

## 4. Formalizing relations in Continuation-Passing Style

The apparent circularity in the definition of the functions corresponding to left-recursive categories suggests that it may be worthwhile reformulating the recognition problem in such a way that the string position results are *produced incrementally*, rather than in one fell swoop, as in the formalization just described. The key insight is that each non-terminal category $A$ in a grammar defines a relation $r_A$ such that $r_A(l,r)$ iff $A$ can derive the substring of the input string spanning string positions $l$ to $r$.[4] Informally speaking, the $r$ can be enumerated one at a time, so the fact that the calculation of $r_A(l,r)$ requires the result $r_A(l,r')$ need not lead to a vicious circularity.

One way to implement this in a functional programming language is to use a 'Continuation-Passing Style' (CPS) of programming.[5] It turns out that a memoized top-down parser writen in continuation-passing style will in fact terminate, even in the face of left recursion. Additionally, the treatment of memoization in a CPS is instructive because it shows the types of table lookup operations needed in chart parsing.

Informally, in a CPS program an additional argument, call it $c$, is added to all functions and procedures. When these functions and procedures are called $c$ is always bound to a procedure (called the continuation); the idea is that a result value $v$ is 'returned' by evaluating ($c$ $v$). For example, the standard definition of the function `square` in (16) would be rewritten in CPS as (17). (18) shows how this definition could be used to compute and display (using the Scheme builtin `display`) the square of the number 3.

(16)
```
(define (square x) (* x x))
```

(17)
```
(define (square cont x) (cont (* x x)))
```

(18)
```
> (square display 3)
9
```

Thus whereas result values in a non-CPS program flow 'upwards' in the procedure call tree, in a CPS program result values flow 'downwards' in the procedure call tree.[6,7] The CPS style of programming can be used to formalize relations in a pure functional language as procedures which can be thought of as 'returning' multiply-valued results any number of times.

These features of CPS can be used to encode CFGs as follows. Each category $A$ is associated with a function $g_A$ that represents the relation $r_A$, i.e., ($g_A$ $c$ $l$) reduces (in an applicative-order reduction) in such a fashion that at some stage in the reduction the expression ($c\ r$) is reduced iff $A$ can derive the substring spanning string positions $l$ to $r$ of the input string. (The value of ($g_A\ c\ l$) is immaterial and therefore unspecified, but see footnote 8 below). That is, if ($g_A\ c\ l$) is evaluated with $l$ bound to the left string position of category $A$, then ($c\ r$) will be evaluated zero or more times with $r$ bound to each of $A$'s right string positions $r$ corresponding to $l$.

---

4 The relation $r_A$ and the function $f_A$ mentioned above satisfy $\forall r\ \forall l\ \ r_A(l,r) \leftrightarrow r \in f(l)$.

5 Several readers of this paper, including a reviewer, suggested that this can be formulated more succinctly using Scheme's `call/cc` continuation-constructing primitive. I confess that I try to avoid this primitive because it is outside of the pure functional subset of Scheme, but in any case neither I nor any of the readers I could challenge could produce a more succinct formulation using `call/cc`.

6 Tail recursion optimization prevents the procedure call stack from growing unboundedly.

7 This CPS formalization of CFGs closely related to the 'downward success passing' method of translating Prolog into Lisp discussed by Kahn and Carlsson (1984).

For example, a CPS function recognizing the terminal item 'will' (arguably a future auxiliary in a class of its own) could be written as (19).

(19)
```
(define (future-aux continuation pos)
  (if (and (pair? pos) (eq? (car pos) 'will))
      (continuation (cdr pos))))
```

For a more complicated example, consider the two rules defining VP in the fragment above, repeated here as (20). These could be formalized as the CPS function defined in (21).

(20) VP → V NP        VP → V S

(21)
```
(define (VP continuation pos)
  (begin
    (V (lambda (pos1) (NP continuation pos1)) pos)
    (V (lambda (pos1) (S continuation pos1)) pos)))
```

In this example `V`, `NP`, and `S` are assumed to have CPS definitions. Informally, the expression `(lambda (pos1) (NP continuation pos1))` is a contuation that specifies what to do if a V is found, viz., pass the V's right string position `pos1` to the NP recognizer as its left-hand string position, and instruct the NP recognizer in turn to pass its right string positions to `continuation`.

The recognition process begins by passing the function corresponding to the root category the string to be recognized, and a continuation (to be evaluated after successful recognition) which records the successful analysis.[8]

[8] Thus this formalization makes use of mutability to return final results, and so cannot be expressed in a purely functional language. However, it is possible to construct a similiar formalization in the purely functional subset of Scheme by passing around an additional 'result' argument (here the last argument). The examples above would be rewritten as the following under this approach.

(19′)
```
(define (future-aux continuation pos result)
  (if (and (pair? pos) (eq? (car pos) 'will))
      (continuation (cdr pos) result)))
```

(21′)
```
(define (VP continuation pos result)
  (V (lambda (pos1 result1)
       (NP continuation pos1 result1))
     pos
     (V (lambda (pos1 result1)
          (S continuation pos1 result1))
        pos
        result)))
```

(22)
```
(define (recognize words)
  (let ((recognized #f))
    (S (lambda (pos)
         (if (null? pos) (set! recognized #t)))
       words)
    recognized))
```

Thus rather than constructing a set of all the right string positions (as in the previous encoding), this encoding exploits the ability of the CPS approach to 'return' a value zero, one or more times (corresponding to the number of right string positions). And although it is not demonstrated in this paper, the ability of a CPS procedure to 'return' more than one value at a time can be used to pass other information besides right string position, such as additional syntactic features or semantic values.

Again, higher-order functions can be used to simply the definitions of the CPS functions corresponding to categories. The CPS versions of the `terminal`, `seq` and `alt` functions are given as (23), (25) and (24) respectively.

(23)
```
(define (terminal word)
  (lambda (continuation pos)
    (if (and (pair? pos) (eq? (car pos) word))
        (continuation (cdr pos)))))
```

(24)
```
(define (alt alt1 alt2)
  (lambda (continuation pos)
    (begin (alt1 continuation pos)
           (alt2 continuation pos))))
```

(25)
```
(define (seq seq1 seq2)
  (lambda (cont pos)
    (seq1 (lambda (pos1) (seq2 cont pos1))
          pos)))
```

If these three functions definitions replace the earlier definitions given in (5), (6) and (7), the fragment in Figure 1 defines a CPS recognizer. Note that just as in the first CFG encoding, the resulting program behaves as a top-down recognizer. Thus in general these programs fail to terminate when faced with a left-recursive grammar for essentially the same reason: the procedures that correspond to left-recursive categories involve ill-founded recursion.

## 5. Memoization in Continuation-Passing Style

The `memo` procedure defined in (15) is not appropriate for CPS programs because it associates the arguments of the functional expression with the value that the expression reduces to, but in a CPS program the 'results' produced by an expression are the values it

---

(22′)
```
(define (recognize words)
  (S (lambda (pos result)
       (if (null? pos) #t result))
     words))
```

passes on to the continuation, rather than the value that the expression reduces to. That is, a memoization procedure for a CPS procedure should associate argument values with the set of values that the unmemoized procedure passes to its continuation. Because an unmemoized CPS procedure can produce multiple result values, its memoized version must store not only these results, but also the continuations passed to it by its callers, which must receive any additional results produced by the original unmemoized procedure.

The `cps-memo` procedure in (26) achieves this by associating a table entry with each set of argument values which has two components; a list of caller continuations and a list of result values. The caller continuation entries are constructed when the memoized procedure is called, and the result values are entered and propagated back to callers each time the unmemoized procedure 'returns' a new value.[9]

(26)
```
(define (memo cps-fn)
  (let ((table (make-table)))
    (lambda (continuation . args)
      (let ((entry (table-ref table args)))
       (cond ((null? (entry-continuations entry))
              ; first time memoized procedure has been called with args
              (push-continuation! entry continuation)
              (apply cps-fn
               (lambda result
                 (when (not (result-subsumed? entry result))
                   (push-result! entry result)
                   (dolist (cont (entry-continuations entry))
                     (apply cont result))))
               args))
             (else
              ; memoized procedure has been called with args before
              (push-continuation! entry continuation)
              (dolist (result (entry-results entry))
               (apply continuation result)))))))))
```

Specifically, when the memoized procedure is called, `continuation` is bound to the continuation passed by the caller that should receive 'return' values, and `args` is bound to a list of arguments that index the entry in the memo table, and are passed to the unmemoized procedure `cps-fn` if evaluation is needed. The memo table `table` initially associates every set of arguments with empty caller continuation and empty result value sets. The local variable `entry` is bound to the table entry that corresponds to `args`; the set of caller continuations stored in `entry` is null iff the memoized function has not been called with this particular set of arguments before.

---

[9] The `dolist` from used in (26) behaves as the `dolist` form in CommonLisp. It can be defined in terms of Scheme primitives as follows:

```
(define-syntax dolist
  (syntax-rules ()
    ((dolist (var list) . body)
     (do ((to-do list))
         ((null? to-do))
       (let ((var (car to-do)))
         . body)))))
```

The `cond` clause determines if the memoized function has been called with `args` before by checking if the continuations component of the table entry is non-empty. In either case, the caller continuation needs to be stored in the continuations component of the table entry, so that it can receive any additional results produced by the unmemoized procedure.

If the memoized procedure has not been called with `args` before, it is necessary to call the unmemoized procedure `cps-fn` to produce the result values for `args`. The continuation passed to `cps-fn` checks to see if each `result` of this evaluation is subsumed by some other result already produced for this entry; if it is not, it is pushed onto the results component of this entry, and finally passed to each caller continuation associated with this entry.

If the memoized procedure has been called with `args` before, the results associated with this table entry can be reused. After storing the caller continuation in the table entry, each result already accumulated in the table entry is passed to the caller continuation.

Efficient implementations of the table and entry manipulation procedures would be specialized for the particular types of arguments and results used by the unmemoized procedures. Here we give a simple and general, but less than optimal, implementation using association lists.[10]

A table is a headed association list (27), which is extended as needed by `table-ref` (28). In this fragment there are no partially specified arguments or results (such as would be involved if the fragment used feature structures), so the subsumption relation is in fact equality.

(27)
```
(define (make-table) (list '*head*))
```

(28).
```
(define (table-ref table key)
  (let ((pair (assoc key (cdr table))))
    (if pair              ; an entry already exists
        (cdr pair)        ; return it
        (let ((new-entry (make-entry)))
          (set-cdr! table (cons (cons key new-entry)
                                (cdr table)))
          new-entry))))
```

Entries are manipulated by the following procedures. Again, because this fragment does not produce partially specified results, the result subsumption check can be performed by the Scheme function `member`.

(29)
```
(define (make-entry) (cons '() '()))
```

(30)
```
(define entry-continuations car)
```

(31)
```
(define entry-results cdr)
```

---

[10] This formalization makes use of 'impure' features of Scheme, specifically destructive assignment to add an element to the `table` list (which is why this list contains the dummy element `*head*`). Arguably, this is a case where impure features result in a more comprehensible overall program.

(32)
```
(define (push-continuation! entry continuation)
  (set-car! entry (cons continuation (car entry))))
```

(33)
```
(define (push-result! entry result)
  (set-cdr! entry (cons result (cdr entry))))
```

(34)
```
(define (result-subsumed? entry result)
  (member result (entry-results entry)))
```

As claimed above, the memoized version of the CPS top-down parser does terminate, even if the grammar is left-recursive. Informally, memoized CPS top-down parsers terminate in the face of left-recursion because they ensure that no unmemoized procedure is ever called twice with the same arguments. For example, we can replace the definition of `NP` in the fragment with the left-recursive one given in (35) without compromising termination, as shown in (36) (where the input string is meant to approximate *Kim's professor knows every student*).

(35)
```
(define NP (memo (vacuous
                  (alt PN              ; NP → PN
                  (alt (seq NP N)      ;    | NP N
                       (seq Det N)))))) ;    | Det N
```

(36)
```
> (recognize '(Kim professor knows every student))
#t
```

Memoized CPS top-down recognizers do in fact correspond fairly closely to chart parsers. Informally, the memo table for the procedure corresponding to a category $A$ will have an entry for an argument string position $l$ just in case a predictive chart parser predicts a category $A$ at position $l$, and that entry will contain string position $r$ as a result just in case the corresponding chart contains a complete edge spanning from $l$ to $r$. Moreover, the evaluation of the procedure $p_A$ corresponding to a category $A$ at string position $l$ corresponds to predicting $A$ at position $l$, and the evaluation of the caller continuations corresponds to the completion steps in chart parsing. The CPS memoization described here caches such evaluations in the same way that the chart caches predictions, and the termination in the face of left recursive follows from the fact that no procedure $p_A$ is ever called with the same arguments twice. Thus given a CPS formalization of the parsing problem and an appropriate memoization technique, it is in fact the case that "the maintenance of well-formed substring tables or charts can be seen as a special case of a more general technique: memoization" (Norvig 1991), even if the grammar contains left-recursion.

## 6. Conclusion and Future Work

This paper has shown how to generalize Norvig's application of memoization to top-down recognizers to yield terminating recognizers for left-recursive grammars. Although not discussed here, the techniques used to construct the CPS recognizers can be generalized to parsers that construct parse trees, or associate categories with "semantic values" or "unifiation-based" feature structures. Specifically, we add extra arguments to each (caller) continuation whose value is the feature structure, parse tree and/or the "semantic value" associated with each category.

Doing this raises other interesting questions not addressed by this paper. As noted by a CL reviewer, while the use of memoization described here achieves termination in the face of left recursion and polynomial recognition times for CFGs, it does not provide packed parse forest representations of the strings analysed in the way that chart-based systems can (Lang 1991, Tomita 1985). Since the information that would be used to construct such packed parse forest representations in a chart is encapsulated inthe state of the memoized functions, a straight-forward implementation attempt would probably be very complicated, and I suspect ultimately not very informative. I suggest that it might be more fruitful to try to develop an appropriate higher level of abstraction. For example, the packed parse forest representation exploits the fact that all that matters about a subtree is its root label and the substring it spans; its other internal details are irrelevant. This observation might be exploited by performing parse tree construction on *streams* of subtrees with the same root labels and string positions (formulated using CPS as described above) rather than individual subtrees; these operations would be 'delayed' until the stream is actually read, as is standard, so the parse trees would not actually be constructed during the parsing process. Whether or not this particular approach is viable is not that important, but it does seem as if a functional perspective provides useful and insightful ways to think about the parsing process.